\documentclass{article}

\usepackage[preprint]{neurips_2026}
\usepackage{graphicx}
\usepackage[utf8]{inputenc} 
\usepackage[T1]{fontenc}    
\usepackage{hyperref}       
\usepackage{url}            
\usepackage{booktabs}       
\usepackage{amsfonts}       
\usepackage{nicefrac}       
\usepackage{microtype}      
\usepackage{xcolor}         
\usepackage{amsmath}   
\usepackage{amssymb}  
\usepackage{booktabs} 
\usepackage{natbib}   
\usepackage{amsmath,amssymb}
\usepackage{algorithm}
\usepackage{algorithmic}
\title{CoBind: Stage-Aware Compositional Binding for Training-Free Text-to-Image Generation}

\author{%
Kaijie Chen \\
Mind Lab \\
\And
Ethan Caldwell \\
Mind Lab \\
\And
Mira Vossen \\
Mind Lab \\
\AND
Julian Hartwell \\
Mind Lab \\
\And
Serena Whitlock \\
Mind Lab \\
\And
Adrian Bellamy \\
Mind Lab \\
}

\begin{document}

\maketitle

\begin{abstract}

Diffusion-based text-to-image models can synthesize visually compelling images, yet they still struggle to faithfully execute complex prompts involving multiple entities, attributes, and relations. Typical failures include object omission, incorrect attribute assignment, and reversed spatial relations. Existing training-free methods mainly improve text-image alignment by strengthening token-level cross-attention. However, a strong response to an individual token does not guarantee that an attribute is assigned to the correct entity, nor does it explicitly enforce the structural relations among multiple objects. Moreover, most existing methods apply a unified guidance objective throughout the denoising process, overlooking the fact that global layouts, entity attributes, and local visual details emerge at different stages of diffusion sampling.

We introduce \textbf{CoBind}, a training-free framework for stage-aware compositional binding. CoBind first represents the input prompt as a composition graph consisting of entities, attributes, and relations, explicitly encoding attribute ownership and inter-entity dependencies. It then applies different structural constraints at different stages of generation. During early denoising steps, CoBind establishes the global scene layout through entity-completeness and relation constraints. During intermediate steps, it binds attributes to their corresponding entities using entity-specific spatial supports and a contrastive cross-entity objective. During late steps, structural guidance is gradually relaxed, allowing the pretrained diffusion model to recover textures and fine-grained visual details. CoBind further adjusts the guidance strength according to the current satisfaction level of each constraint, thereby avoiding unnecessary perturbations once a constraint has already been fulfilled.

CoBind modifies only the latent variables during sampling and requires neither model retraining nor additional image annotations. We evaluate CoBind on T2I-CompBench++, GenEval, and multiple pretrained text-to-image diffusion models. The results show consistent improvements in attribute binding, inter-entity relations, and complex compositional generation, while preserving competitive visual quality. Ablation studies further validate the individual contributions of composition-graph modeling, stage-aware scheduling, contrastive attribute binding, and satisfaction-adaptive optimization.

\end{abstract}
\section{Introduction}
\label{sec:intro}

Text-to-image generation models aim to transform natural-language descriptions into realistic and semantically consistent visual content. In recent years, diffusion models have substantially improved image resolution, visual quality, and open-domain generalization by learning complex image distributions through iterative denoising and conditioning the generation process on text~\citep{ho2020denoising,rombach2022high,podell2023sdxl}. Latent diffusion models, in particular, move the generative process from pixel space to a compressed latent representation, reducing training and inference costs while preserving visual details. This design enables large-scale text-to-image systems to handle a broad range of objects, scenes, and artistic styles. For structurally simple descriptions such as ``a photo of a red sports car on a snowy road,'' existing models can usually produce images with high visual quality that broadly conform to the textual semantics.

However, improved image quality does not necessarily imply faithful execution of the text. When a prompt simultaneously contains multiple entities, entity attributes, and inter-entity relations, current models still exhibit systematic compositional errors. For example, given ``a red cube to the left of a blue sphere,'' a model may omit one of the objects, bind a color to the wrong entity, or generate both objects while reversing their spatial positions. These errors accumulate as the number of entities and constraints in the prompt increases. Existing compositional text-to-image benchmarks evaluate such capabilities along dimensions including color, shape, and texture binding; spatial and non-spatial relations; and complex multi-object compositions. Their results show that even high-quality text-to-image models remain noticeably unstable across these dimensions~\citep{huang2023t2icompbench,huang2025t2icompbench++}. Therefore, the central challenge facing current text-to-image models is no longer merely whether they can generate realistic images, but whether they can reliably translate an internally structured textual description into the corresponding visual structure.

Compositional errors are not simply failures of object recognition. A complex prompt typically specifies information at three levels: which entities should appear in the image, which attributes each entity should possess, and which relations should hold among those entities. Existing diffusion models generally encode the complete prompt into a set of textual representations and inject them into shared visual features through cross-attention. Consequently, different textual components compete for limited attention and spatial regions within the same generative space. When multiple entities are semantically similar, or when multiple attributes must be bound separately to different entities, the model lacks an explicit mechanism for preserving the ownership relations among textual components. Meanwhile, diffusion sampling exhibits a clear temporal structure: early denoising steps largely determine the global layout and object positions, intermediate steps progressively establish entity boundaries and semantic attributes, and late steps add textures and local details. If the spatial structure is incorrectly determined at an early stage, subsequent steps usually struggle to rearrange the entities. Likewise, if attributes become entangled during the intermediate stage, merely strengthening the attention of the corresponding tokens may not recover the correct bindings.

Existing methods primarily improve compositional text--image consistency along three directions. The first line of work improves overall generation capability by scaling training data, model capacity, or text encoders. Although these approaches can improve average generation quality, they require expensive retraining and do not directly constrain the correspondence structure among entities, attributes, and relations. The second line of work optimizes cross-attention at inference time. For example, Attend-and-Excite strengthens the attention responses of subject tokens to reduce the probability of omitting important entities~\citep{chefer2023attend}, while methods such as Divide-and-Bind further attempt to separate the attention regions of different entities or attributes~\citep{li2023divide}. These methods require no parameter updates, but typically treat token-attention strength or spatial separation as a unified optimization objective. A strong token response may indicate that the model attends to a concept, but it does not ensure that an attribute is assigned to the correct entity, nor can it directly represent relations such as `to the left of,'' `holding,'' or ``inside.'' The third line of work decomposes prompts into sub-prompts, scene layouts, or structured representations and then controls image generation through external modules. These methods provide more explicit structural supervision, but often rely on additional language models, detectors, or layout annotations. Parsing errors can propagate into subsequent generation stages, while the additional components increase system complexity and inference overhead.

These approaches share an important limitation: they typically apply the same compositional constraints throughout all denoising steps, overlooking the fact that different semantic errors emerge at different stages of generation. Spatial relations must be established before the global layout becomes fixed; attribute binding must be constrained while entity regions are taking shape; and visual details should be generated freely once the semantic structure has stabilized. Strengthening local attributes too early may disrupt the global layout, whereas correcting spatial relations too late may fail to alter an already established image structure. This leads to a more specific question: \emph{Can compositional constraints be applied at the denoising stages in which the corresponding semantics are most readily determined, according to the internal structure of the prompt, rather than through a single guidance objective that remains active throughout the entire sampling process?}

Motivated by this observation, we propose \textbf{CoBind}, a stage-aware semantic binding framework for compositional text-to-image generation. CoBind first represents the input prompt as a compositional graph consisting of entities, attributes, and relations. Attribute nodes are explicitly connected to their associated entities, while relation edges describe spatial or semantic constraints between entities. CoBind then divides the denoising process into three stages---layout establishment, entity binding, and detail refinement---according to the formation state of visual structures during diffusion. During layout establishment, the method uses inter-entity relations to constrain the relative positions of the corresponding attention regions, prioritizing the global scene structure. During entity binding, it jointly uses cross-attention and visual-region responses to restrict each attribute to the spatial support of its associated entity. During detail refinement, it gradually weakens structural guidance, allowing the pretrained model to recover textures, lighting, and local visual details. Unlike approaches that apply the same optimization to all tokens, CoBind dynamically determines which guidance objectives should be activated according to the semantic-node type, the current denoising stage, and the degree to which each constraint has been satisfied.

CoBind is designed as a training-free inference-time module that can be directly integrated into existing latent diffusion models. Its core objective is not to continuously increase the attention assigned to all text tokens, but to preserve the correspondences among entities, attributes, and relations throughout generation. The compositional graph specifies the semantic dependencies that must be satisfied, stage-aware scheduling determines when each type of dependency should intervene, and adaptive constraints prevent already satisfied conditions from being repeatedly reinforced. In this way, the model can execute structurally complex textual instructions more reliably while preserving its original generative prior and visual quality.

Our main contributions are as follows:

\begin{itemize}
\item We analyze compositional text-to-image errors from the temporal perspective of diffusion sampling and identify distinct effective intervention stages for different types of compositional constraints: entity relations primarily depend on early-stage layout formation, attribute bindings are mainly established during intermediate stages, and visual details should remain only weakly constrained during late stages.
 
\item We propose CoBind, which models a prompt as a compositional graph of entities, attributes, and relations and employs a stage-aware semantic binding mechanism to constrain spatial relations, entity completeness, and attribute ownership at their corresponding denoising stages.

\item CoBind is a general training-free inference-time framework that can be applied to different pretrained text-to-image diffusion models without additional image annotations or parameter updates.

\item We conduct systematic evaluations on T2I-CompBench++, GenEval, and multiple pretrained text-to-image models. The results show that CoBind consistently improves attribute binding, object relations, and complex composition while maintaining competitive visual quality. The exact improvements will be added after the experiments are completed.

\end{itemize}

\begin{figure}[t]
    \centering
    \includegraphics[width=\linewidth]{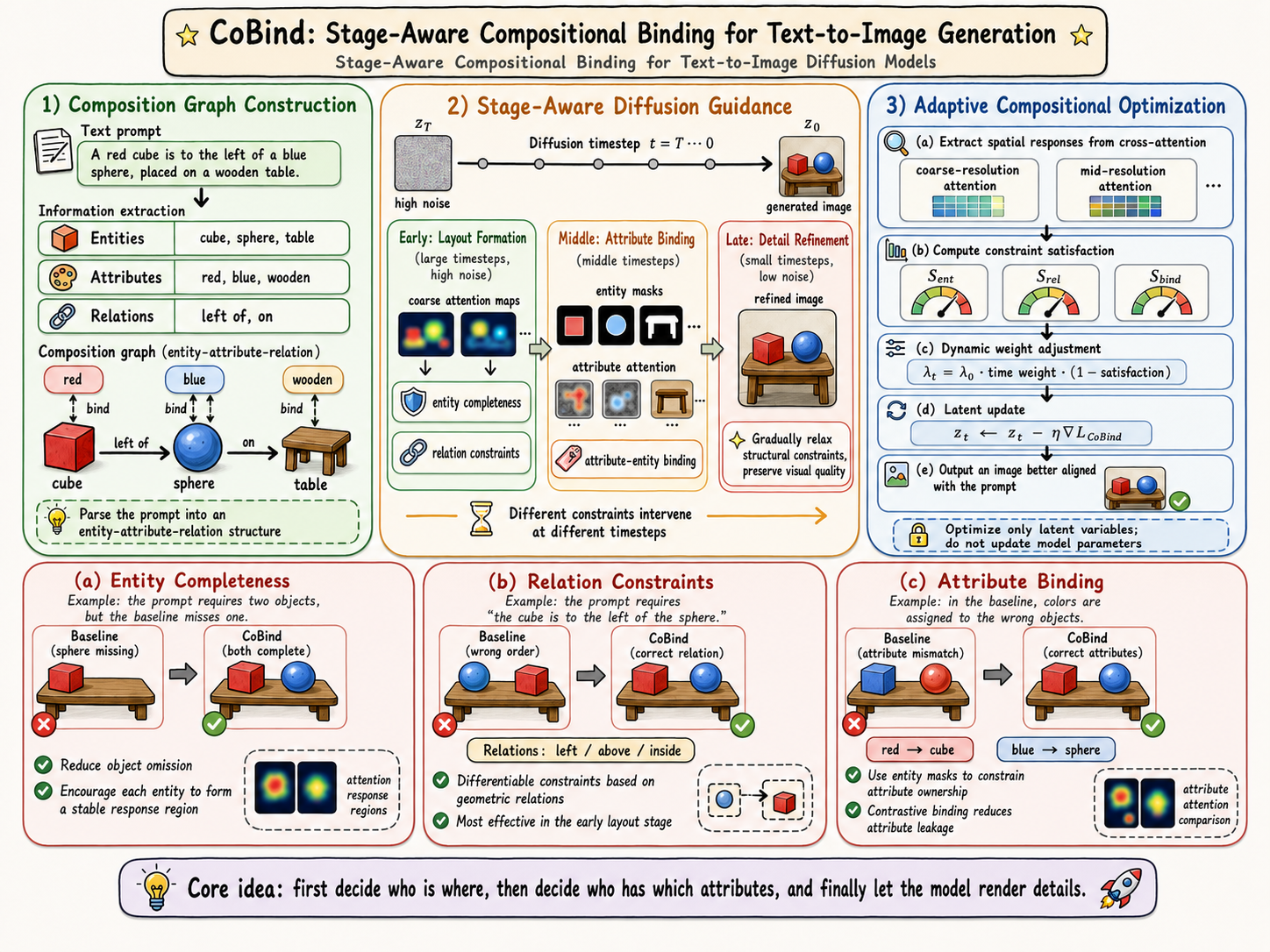}
    \caption{Overview of CoBind. Given a text prompt, CoBind constructs an entity--attribute--relation composition graph and applies stage-aware constraints during diffusion sampling. Layout and relations are established at early timesteps, attributes are bound to their corresponding entities at intermediate timesteps, and structural guidance is gradually relaxed during late-stage detail refinement.}
    \label{fig:method}
\end{figure}

\section{Method}
\label{sec:method}

\subsection{Problem Formulation and Method Overview}
\label{sec:method_overview}

Given a text prompt $P$, the objective of a text-to-image diffusion model is to generate an image $x$ that achieves high visual quality while satisfying the entity, attribute, and relation constraints specified in the prompt. We consider a pretrained latent diffusion model whose denoising network is denoted by
\begin{equation}
\epsilon_{\theta}(z_t,t,c),
\end{equation}
where $z_t$ denotes the noisy latent variable at diffusion timestep $t$, $c=\mathcal{E}(P)$ is the conditional representation produced by the text encoder $\mathcal{E}$, and $\theta$ denotes the frozen model parameters.

Existing text-to-image models typically encode the entire prompt into a sequence of text tokens and inject the textual information into visual features through cross-attention. This mechanism can capture the overall semantic correspondence between the image and the text, but it does not explicitly preserve the following three types of compositional constraints:

\begin{enumerate}
\item which independent entities should appear in the prompt;
\item which entity each attribute specifically belongs to;
\item which spatial or interaction relations should hold between different entities.
\end{enumerate}

Consequently, even when the model recognizes every concept in the prompt individually, it may still combine these concepts incorrectly. For example, the model may generate ``red,'' ``cube,'' ``blue,'' and ``sphere'' simultaneously, but assign blue to the cube and red to the sphere. In this case, all individual concepts are present, yet the compositional structure remains incorrect.

To address this issue, we propose \textbf{CoBind}, a stage-aware compositional binding framework. Given a prompt, CoBind first converts it into an explicit compositional graph that represents the dependency structure among entities, attributes, and relations. CoBind then extracts the spatial responses associated with individual semantic nodes from the cross-attention maps of the diffusion model and applies different types of constraints at different denoising stages:

\begin{itemize}
\item During early denoising, CoBind prioritizes determining whether the required entities are present and establishing their relative layout;
\item During intermediate denoising, it binds attributes such as color, material, and shape to their corresponding entities;
\item During late denoising, it gradually weakens the structural constraints and returns generation freedom to the pretrained model, allowing it to recover textures and local details.
\end{itemize}

CoBind does not update the parameters of the diffusion model. Instead, during sampling, it applies a small number of gradient-based corrections to the current latent variable $z_t$ according to the compositional constraints. The overall method can be written as
\begin{equation}
z_t^{\star}
=
z_t-\eta_t\nabla_{z_t}\mathcal{L}_{\mathrm{CoBind}}^{t},
\end{equation}
where $\eta_t$ denotes the correction step size at timestep $t$, and $\mathcal{L}_{\mathrm{CoBind}}^{t}$ is the guidance objective jointly determined by the current stage, the compositional graph structure, and the constraint-satisfaction state.

\subsection{Prompt Compositional Graph}
\label{sec:composition_graph}

\subsubsection{Graph Definition}

We convert the prompt $P$ into a compositional graph
\begin{equation}
\mathcal{G}
=
\left(
\mathcal{V}_{E},
\mathcal{V}_{A},
\mathcal{E}_{B},
\mathcal{E}_{R}
\right),
\end{equation}
where $\mathcal{V}_{E}$ denotes the set of entity nodes, $\mathcal{V}_{A}$ denotes the set of attribute nodes, $\mathcal{E}_{B}$ denotes the set of attribute-binding edges, and $\mathcal{E}_{R}$ denotes the set of inter-entity relation edges.

Each entity node $e_i\in\mathcal{V}_{E}$ corresponds to an independent visual object in the prompt, such as ``cube,'' ``sphere,'' or ``table.'' Each attribute node $a_j\in\mathcal{V}_{A}$ corresponds to a color, material, shape, quantity, or texture description, such as ``red,'' ``wooden,'' or ``striped.''

An attribute-binding edge
\begin{equation}
(a_j,e_i)\in\mathcal{E}_{B}
\end{equation}
indicates that attribute $a_j$ should belong to entity $e_i$. A relation edge is represented as
\begin{equation}
(e_i,r_{ij},e_j)\in\mathcal{E}_{R},
\end{equation}
where $r_{ij}$ denotes the type of relation between the two entities, such as \textit{left of}, \textit{above}, \textit{inside}, \textit{near}, or \textit{on}.

Consider the prompt

\begin{quote}
``a red cube to the left of a blue sphere on a wooden table''
\end{quote}

as an example. The compositional graph contains three entity nodes:
\begin{equation}
\mathcal{V}_{E}
=
\{\text{cube},\text{sphere},\text{table}\},
\end{equation}
three attribute nodes:
\begin{equation}
\mathcal{V}_{A}
=
\{\text{red},\text{blue},\text{wooden}\},
\end{equation}
and the attribute-binding relations
\begin{equation}
\begin{split}
\text{red}&\rightarrow\text{cube},\\
\text{blue}&\rightarrow\text{sphere},\\
\text{wooden}&\rightarrow\text{table}.
\end{split}
\end{equation}

The compositional graph also contains the spatial relation
\begin{equation}
(\text{cube},\text{left of},\text{sphere}),
\end{equation}
as well as the support relation extracted from the prepositional structure.

\subsubsection{Compositional Graph Construction}

CoBind constructs the compositional graph through a lightweight syntactic parsing procedure. Specifically, we first extract nouns and noun phrases from the prompt as candidate entities. We then identify attributes and their associated entities according to adjectival modification, noun-compound structures, and prepositional dependencies. Finally, we extract inter-entity relations from prepositions, directional terms, and relational verbs.

The compositional graph is constructed only once before generation begins. It requires neither image annotations nor the training of an additional prompt-parsing model. For each graph node, we record the set of corresponding tokens in the tokenized prompt. When an entity or attribute consists of multiple subtokens, we average their attention maps to obtain the overall spatial response of the semantic node.

\subsection{Extracting Spatial Responses from Cross-Attention}
\label{sec:attention_maps}

At the $l$-th cross-attention layer of the diffusion model, the attention between visual-feature queries and textual keys can be written as
\begin{equation}
\mathbf{A}_{t}^{l}
=
\operatorname{Softmax}
\left(
\frac{
\mathbf{Q}_{t}^{l}
(\mathbf{K}^{l}_{c})^{\top}
}{\sqrt{d}}
\right),
\end{equation}
where $\mathbf{Q}_{t}^{l}$ is obtained from the visual features corresponding to the current latent variable, $\mathbf{K}^{l}_{c}$ is obtained from the text condition, and $d$ denotes the dimensionality of the attention features.

For a node $v$ in the compositional graph, let $\mathcal{T}(v)$ denote its corresponding set of text tokens. We aggregate the responses across different tokens and attention heads to obtain the spatial attention map of node $v$ at layer $l$:
\begin{equation}
A_{v,t}^{l}
=
\frac{1}{|\mathcal{T}(v)|}
\sum_{k\in\mathcal{T}(v)}
\operatorname{MeanHead}
\left(
\mathbf{A}_{t}^{l}[:,:,k]
\right).
\end{equation}

Attention layers at different resolutions play different roles in diffusion models. Low-resolution features are better suited to representing global positions and scene layouts, whereas higher-resolution features provide more precise entity boundaries. We therefore construct separate layout and binding attention maps:
\begin{equation}
A_{v,t}^{\mathrm{layout}}
=
\sum_{l\in\mathcal{L}_{\mathrm{coarse}}}
\omega_l A_{v,t}^{l},
\end{equation}
\begin{equation}
A_{v,t}^{\mathrm{bind}}
=
\sum_{l\in\mathcal{L}_{\mathrm{middle}}}
\omega_l A_{v,t}^{l},
\end{equation}
where $\mathcal{L}_{\mathrm{coarse}}$ denotes the set of low-resolution attention layers, $\mathcal{L}_{\mathrm{middle}}$ denotes the set of intermediate-resolution attention layers, and $\omega_l$ is a normalized layer weight.

This resolution separation is important. Directly constraining entity positions on high-resolution attention maps may produce unstable spatial gradients because of local texture noise. Conversely, using only low-resolution attention maps for attribute binding may cause the spatial regions of different entities to overlap because the resolution is insufficient.

\subsection{Stage-Aware Compositional Constraints}
\label{sec:stage_aware_guidance}

Diffusion generation gradually transitions from a highly noisy state to a clear image. Let
\begin{equation}
p_t=1-\frac{t}{T}
\end{equation}
denote the generation progress, where $p_t=0$ corresponds to the beginning of sampling and $p_t=1$ corresponds to the end of sampling.

Rather than optimizing all compositional constraints simultaneously at every timestep, CoBind defines different temporal activation windows for different constraints. Let $\rho_1$ and $\rho_2$ denote the soft boundaries of the layout and binding stages, respectively, where
\begin{equation}
0<\rho_1<\rho_2<1.
\end{equation}

The base temporal weight of the layout constraints is defined as
\begin{equation}
w_{\mathrm{layout}}(p_t)
=
1-
\sigma
\left(
\frac{p_t-\rho_1}{\tau_s}
\right),
\end{equation}
and the temporal weight of the attribute-binding constraints is defined as
\begin{equation}
\begin{split}
w_{\mathrm{bind}}(p_t)
=
&\sigma
\left(
\frac{p_t-\rho_1}{\tau_s}
\right)\\
&\cdot
\left[
1-
\sigma
\left(
\frac{p_t-\rho_2}{\tau_s}
\right)
\right],
\end{split}
\end{equation}
where $\sigma(\cdot)$ is the sigmoid function, and $\tau_s$ controls the smoothness of the transition between stages.

This soft schedule avoids abruptly switching optimization objectives at a particular timestep. Layout guidance maintains a high weight during early sampling and gradually weakens as the global structure emerges. Attribute binding is activated after the entity regions begin to stabilize and gradually deactivates before late-stage detail generation.

\subsection{Entity Completeness Constraint}
\label{sec:entity_presence}

The first type of failure in compositional generation is entity omission. For an entity node $e_i$, using only the maximum value of its attention map as evidence of presence can be unstable, because the model may produce a high response at a single spatial location without generating a complete entity.

We therefore average the $K$ highest-response locations in the entity attention map:
\begin{equation}
q_{\mathrm{ent}}(e_i,t)
=
\operatorname{TopKMean}
\left(
A_{e_i,t}^{\mathrm{layout}},K
\right).
\end{equation}

The entity-completeness loss is defined as
\begin{equation}
\mathcal{L}_{\mathrm{ent}}^{t}
=
\frac{1}{|\mathcal{V}_{E}|}
\sum_{e_i\in\mathcal{V}_{E}}
\left[
1-q_{\mathrm{ent}}(e_i,t)
\right].
\end{equation}

This objective encourages each entity to form a continuous and stable attention response within a local region, rather than merely increasing the attention peak at a single pixel location.

When a prompt contains multiple entities, different entities may compete for the same visual region. For entity pairs whose semantics do not imply overlap, we further define an entity-separation loss:
\begin{equation}
\mathcal{L}_{\mathrm{sep}}^{t}
=
\frac{1}{|\mathcal{P}|}
\sum_{(e_i,e_j)\in\mathcal{P}}
\operatorname{Dice}
\left(
A_{e_i,t}^{\mathrm{layout}},
A_{e_j,t}^{\mathrm{layout}}
\right),
\end{equation}
where $\mathcal{P}$ denotes the set of entity pairs that should occupy distinct spatial regions, and $\operatorname{Dice}(\cdot,\cdot)$ measures the overlap between two attention maps.

For relations such as ``person wearing a hat'' or ``object inside a box,'' which permit or require partial spatial overlap, we do not apply a uniform separation constraint. Instead, the corresponding relation objective determines the appropriate form of overlap.

\subsection{Inter-Entity Relation Constraints}
\label{sec:relation_constraint}

For each entity $e_i$, we normalize its layout attention map into a spatial probability distribution:
\begin{equation}
\widehat{A}_{e_i,t}(u)
=
\frac{
A_{e_i,t}^{\mathrm{layout}}(u)
}{
\sum_{u'}A_{e_i,t}^{\mathrm{layout}}(u')+\varepsilon
},
\end{equation}
where $u=(x,y)$ denotes a spatial location.

The soft centroid of the entity is defined as
\begin{equation}
\boldsymbol{\mu}_{e_i,t}
=
\sum_u u\widehat{A}_{e_i,t}(u).
\end{equation}

For the ``left of'' relation, the relation loss is defined as
\begin{equation}
\mathcal{L}_{\mathrm{left}}^{t}(e_i,e_j)
=
\max
\left(
0,
\mu_{e_i,t}^{x}
-
\mu_{e_j,t}^{x}
+
m_x
\right),
\end{equation}
where $m_x$ is the minimum horizontal margin.

Similarly, the ``above'' relation is represented by
\begin{equation}
\mathcal{L}_{\mathrm{above}}^{t}(e_i,e_j)
=
\max
\left(
0,
\mu_{e_i,t}^{y}
-
\mu_{e_j,t}^{y}
+
m_y
\right).
\end{equation}

For the ``near'' relation, we constrain the distance between the two entity centroids not to exceed the threshold $d_{\mathrm{near}}$:
\begin{equation}
\mathcal{L}_{\mathrm{near}}^{t}(e_i,e_j)
=
\max
\left(
0,
|
\boldsymbol{\mu}_{e_i,t}
-
\boldsymbol{\mu}_{e_j,t}
|*2
-
d_{\mathrm{near}}
\right).
\end{equation}

For the ``far from'' relation, we instead use the opposite margin constraint:
\begin{equation}
\mathcal{L}_{\mathrm{far}}^{t}(e_i,e_j)
=
\max
\left(
0,
d_{\mathrm{far}}
-
|
\boldsymbol{\mu}_{e_i,t}
-
\boldsymbol{\mu}_{e_j,t}
|_2
\right).
\end{equation}

For the ``inside'' relation, suppose that entity $e_i$ should lie inside entity $e_j$. We construct a soft spatial support region $M_{e_j,t}$ from the attention response of $e_j$ and define the containment loss as
\begin{equation}
\mathcal{L}_{\mathrm{inside}}^{t}(e_i,e_j)
=
1-
\frac{
\sum_u
A_{e_i,t}^{\mathrm{layout}}(u)
M_{e_j,t}(u)
}{
\sum_u
A_{e_i,t}^{\mathrm{layout}}(u)
+
\varepsilon
}.
\end{equation}

All relation constraints can be written uniformly as
\begin{equation}
\mathcal{L}_{\mathrm{rel}}^{t}
=
\frac{1}{|\mathcal{E}_{R}|}
\sum_{(e_i,r,e_j)\in\mathcal{E}_{R}}
\psi_r
\left(
A_{e_i,t}^{\mathrm{layout}},
A_{e_j,t}^{\mathrm{layout}}
\right),
\end{equation}
where $\psi_r$ is a differentiable spatial constraint function associated with relation $r$.

Importantly, CoBind does not require every relation to be reduced to centroid ordering. Centroid-based constraints are appropriate for directional relations such as ``left of'' and ``above,'' whereas region overlap, containment ratios, and boundary distances are used to represent relations such as ``inside,'' ``on,'' and contact relations. Different relations employ different geometric definitions, avoiding the inadequate treatment of every relation as a global positional ordering problem.

\subsection{Attribute--Entity Binding}
\label{sec:attribute_binding}

Even when entities appear at the correct positions, their attributes may still be bound to the wrong objects. To address this problem, CoBind constructs a soft spatial support for each entity from its attention map:
\begin{equation}
M_{e_i,t}(u)
=
\sigma
\left(
\frac{
A_{e_i,t}^{\mathrm{bind}}(u)-\beta_{e_i,t}
}{\tau_m}
\right),
\end{equation}
where $\beta_{e_i,t}$ is a threshold adaptively determined from the current attention distribution, and $\tau_m$ controls the smoothness of the soft-mask boundary.

When computing the attribute-binding gradient, we apply a stop-gradient operation to the entity support region:
\begin{equation}
\widetilde{M}_{e_i,t}
=
\operatorname{sg}
\left(
M_{e_i,t}
\right).
\end{equation}

This prevents attribute optimization from moving the entity region in the backward pass, thereby establishing a relatively stable hierarchy between entity layout and attribute binding.

For an attribute node $a_j$ and its target entity $e_i$, we define their spatial matching score as
\begin{equation}
s(a_j,e_i,t)
=
\frac{
\sum_u
A_{a_j,t}^{\mathrm{bind}}(u)
\widetilde{M}_{e_i,t}(u)
}{
\sum_u
A_{a_j,t}^{\mathrm{bind}}(u)
+
\varepsilon
}.
\end{equation}

This score measures the proportion of the attribute attention that falls within the spatial region of the target entity. Maximizing only the matching score for the target entity is still insufficient, because the same attribute may simultaneously diffuse across multiple entities. We therefore employ a cross-entity contrastive objective:
\begin{equation}
\mathcal{L}_{\mathrm{bind}}^{t}(a_j,e_i)
=
-
\log
\frac{
\exp\left(s(a_j,e_i,t)/\tau_b\right)
}{
\sum_{e_k\in\mathcal{V}_{E}}
\exp\left(s(a_j,e_k,t)/\tau_b\right)
},
\end{equation}
where $\tau_b$ is the contrastive temperature.

The overall attribute-binding loss is
\begin{equation}
\mathcal{L}_{\mathrm{bind}}^{t}
=
\frac{1}{|\mathcal{E}_{B}|}
\sum_{(a_j,e_i)\in\mathcal{E}_{B}}
\mathcal{L}_{\mathrm{bind}}^{t}(a_j,e_i).
\end{equation}

This objective performs two operations simultaneously: it pulls the attribute response toward its associated entity and pushes it away from competing entities. For example, given ``a red cube and a blue sphere,'' the red attention must not only cover the cube but also avoid covering the sphere. Compared with independently increasing the maximum attention of the ``red'' token, cross-entity contrastive learning more directly constrains attribute ownership.

\subsection{Constraint-Satisfaction-Driven Adaptive Guidance}
\label{sec:adaptive_guidance}

A fixed temporal schedule still has one limitation: different prompts have different levels of generation difficulty. A simple relation may already be satisfied at an early stage, whereas a more complex relation may remain incorrect at the same timestep. Continuing to optimize constraints that have already been satisfied may excessively separate object positions, distort entity shapes, or degrade image quality.

CoBind therefore estimates the satisfaction level of each constraint in addition to applying temporal weights.

The entity-satisfaction score is defined as
\begin{equation}
S_{\mathrm{ent}}^{t}
=
\frac{1}{|\mathcal{V}_{E}|}
\sum_{e_i\in\mathcal{V}_{E}}
q_{\mathrm{ent}}(e_i,t).
\end{equation}

The attribute-binding satisfaction score is defined as
\begin{equation}
S_{\mathrm{bind}}^{t}
=
\frac{1}{|\mathcal{E}_{B}|}
\sum_{(a_j,e_i)\in\mathcal{E}_{B}}
s(a_j,e_i,t).
\end{equation}

For relation constraints, we convert the relation loss into a satisfaction score:
\begin{equation}
S_{\mathrm{rel}}^{t}
=
\frac{1}{|\mathcal{E}_{R}|}
\sum_{(e_i,r,e_j)\in\mathcal{E}_{R}}
\exp
\left(
-\psi_r(e_i,e_j)
\right).
\end{equation}

The final dynamic constraint weights are defined as
\begin{equation}
\lambda_{\mathrm{rel}}^{t}
=
\lambda_{\mathrm{rel}}^{0}
w_{\mathrm{layout}}(p_t)
\left(
1-\operatorname{sg}
\left(
S_{\mathrm{rel}}^{t}
\right)
\right),
\end{equation}
\begin{equation}
\lambda_{\mathrm{bind}}^{t}
=
\lambda_{\mathrm{bind}}^{0}
w_{\mathrm{bind}}(p_t)
\left(
1-\operatorname{sg}
\left(
S_{\mathrm{bind}}^{t}
\right)
\right).
\end{equation}

When a constraint remains unsatisfied, its guidance weight is high. Once the constraint has been stably satisfied, its weight automatically decays. The stop-gradient operation ensures that the satisfaction score is used only to regulate optimization strength and does not introduce an additional gradient path through which the model could opportunistically reduce the loss.

This design extends CoBind from a fixed stage schedule to a dynamic guidance mechanism that is both stage-aware and state-aware. The temporal stage determines whether a particular type of constraint is suitable for intervention, while the current satisfaction level determines whether that constraint still requires further intervention.

\subsection{Combined Guidance Objective and Latent Update}
\label{sec:combined_objective}

At timestep $t$, the complete CoBind guidance objective is
\begin{equation}
\begin{split}
\mathcal{L}_{\mathrm{CoBind}}^{t}
=
&\lambda_{\mathrm{ent}}^{t}
\mathcal{L}_{\mathrm{ent}}^{t}
+
\lambda_{\mathrm{sep}}^{t}
\mathcal{L}_{\mathrm{sep}}^{t}\\
&+
\lambda_{\mathrm{rel}}^{t}
\mathcal{L}_{\mathrm{rel}}^{t}
+
\lambda_{\mathrm{bind}}^{t}
\mathcal{L}_{\mathrm{bind}}^{t}
+
\lambda_{\mathrm{reg}}
\mathcal{L}_{\mathrm{reg}}^{t}.
\end{split}
\end{equation}

To limit the deviation of the guided process from the original generation trajectory, we introduce a latent trust-region regularizer. Let $z_t^{(0)}$ denote the latent variable before compositional optimization at the current timestep. We define
\begin{equation}
\mathcal{L}_{\mathrm{reg}}^{t}
=
\left|
z_t-z_t^{(0)}
\right|_2^2.
\end{equation}

At each activated timestep, we perform $K_t$ local updates:
\begin{equation}
z_t^{(k+1)}
=
z_t^{(k)} -
\eta_t
\operatorname{Clip}
\left(
\nabla_{z_t^{(k)}}
\mathcal{L}_{\mathrm{CoBind}}^{t},
g_{\max}
\right),
\end{equation}
where $g_{\max}$ denotes the gradient-clipping threshold. One or a small number of updates can be performed during early and intermediate timesteps. After entering the detail-refinement stage, we set $K_t=0$ and fully restore the original diffusion sampling process.

After the compositional correction is completed, the original diffusion scheduler computes the next timestep:
\begin{equation}
z_{t-1}
=
\operatorname{Step}
\left(
z_t^{\star},
\epsilon_{\theta}
\left(
z_t^{\star},t,c
\right)
\right).
\end{equation}

CoBind does not update $\theta$ and does not require retraining either the text encoder or the diffusion model. Gradients are applied only to the latent variable at the current timestep.

\subsection{Complete Inference Pipeline}
\label{sec:inference_pipeline}

The complete CoBind generation procedure is as follows:

\begin{enumerate}
\item Syntactically parse the input prompt $P$ and construct a compositional graph $\mathcal{G}$ containing entities, attribute bindings, and inter-entity relations.

\item Obtain the text condition $c$ using the pretrained text encoder and initialize the noisy latent variable $z_T$ from a Gaussian distribution.

\item At each denoising timestep $t=T,\ldots,1$, run the frozen diffusion network and collect the cross-attention maps.

\item Extract entity-layout responses from low-resolution attention layers and attribute-binding responses from intermediate-resolution attention layers.

\item Compute the stage weights for the layout and attribute-binding constraints according to the current generation progress.

\item Estimate the current satisfaction levels of the different constraints according to entity completeness, relation correctness, and attribute-binding status.

\item Apply gradient-based corrections to the latent variable for compositional constraints that remain unsatisfied and are still active at the current stage.

\item Perform the standard denoising update using the corrected latent variable. Stop compositional correction during late timesteps and allow the original diffusion model to generate textures and details.

\item After denoising is complete, decode the final latent variable $z_0$ into the image $x$ using the pretrained decoder.

\end{enumerate}

The algorithm can be summarized as follows:

\begin{algorithm}[t]
\caption{CoBind Stage-Aware Compositional Generation}
\label{alg:cobind}
\begin{algorithmic}[1]
\REQUIRE Text prompt $P$, frozen diffusion model $\epsilon_{\theta}$, number of timesteps $T$
\ENSURE Generated image $x$
\STATE Construct the compositional graph $\mathcal{G}\leftarrow\operatorname{Parse}(P)$
\STATE Compute the text condition $c\leftarrow\mathcal{E}(P)$
\STATE Initialize $z_T\sim\mathcal{N}(0,I)$
\FOR{$t=T,\ldots,1$}
\STATE Set $z_t^{(0)}\leftarrow z_t$
\STATE Compute the generation progress $p_t=1-t/T$
\STATE Extract node attention maps $\{A_{v,t} \mid v\in\mathcal{G}\}$
\STATE Compute the constraint-satisfaction scores $S_{\mathrm{ent}}^{t}$, $S_{\mathrm{rel}}^{t}$, and $S_{\mathrm{bind}}^{t}$
\STATE Compute the stage weights and dynamic constraint weights
\FOR{$k=0,\ldots,K_t-1$}
\STATE Compute $\mathcal{L}_{\mathrm{CoBind}}^{t}$
\STATE $z_t^{(k+1)}
\leftarrow
z_t^{(k)}
-
\eta_t\nabla_{z_t^{(k)}}
\mathcal{L}_{\mathrm{CoBind}}^{t}$
\ENDFOR
\STATE Perform standard diffusion denoising using $z_t^{\star}=z_t^{(K_t)}$
\STATE Obtain $z_{t-1}$
\ENDFOR
\STATE $x\leftarrow\operatorname{Decode}(z_0)$
\RETURN $x$
\end{algorithmic}
\end{algorithm}

\subsection{Discussion}
\label{sec:method_discussion}

CoBind differs from conventional attention-enhancement methods in three key respects.

First, CoBind does not optimize the attention strength of isolated tokens. Instead, it optimizes the structural relations represented in the compositional graph. Whether an attribute is correctly generated depends not only on whether its own attention is sufficiently strong, but also on whether it falls within the region of the correct entity and maintains sufficiently low matching scores with other entities.

Second, CoBind distinguishes the temporal stages and feature resolutions required for global layout and local binding. Relation constraints mainly operate on low-resolution attention during early sampling, whereas attribute binding mainly operates on higher-resolution attention during intermediate sampling. This avoids using the same attention layers and the same temporal window to address every type of compositional error.

Third, CoBind dynamically terminates guidance that is no longer necessary according to the degree of constraint satisfaction. It does not continuously move entities or strengthen attributes merely to reduce a fixed loss, thereby reducing the conflict between compositional consistency and visual quality.

The core of CoBind is therefore not simply to increase attention responses, but to maintain a visual compositional structure that is progressively instantiated during diffusion sampling: the early stage determines ``which entities are located where,'' the intermediate stage determines ``which attributes belong to which entities,'' and the late stage allows the pretrained model to determine ``what visual details these entities should ultimately exhibit.''

\section{Experiments}
\label{sec:experiments}

We conduct experiments to answer the following four questions:

\begin{enumerate}
\item Can CoBind improve the text consistency of text-to-image models on attribute binding, entity relations, and complex compositional tasks?
\item Does stage-aware scheduling outperform uniformly applying compositional constraints at all denoising timesteps?
\item How much performance gain is contributed by the composition graph, contrastive attribute binding, and satisfaction-aware adaptation, respectively?
\item Can CoBind preserve the visual quality and generation diversity of the original model while improving compositional correctness?
\end{enumerate}

\subsection{Experimental Setup}
\label{sec:experimental_setup}

\subsubsection{Evaluation Benchmarks}

\paragraph{T2I-CompBench++.}

We first evaluate compositional generation ability on T2I-CompBench++~\citep{huang2025t2icompbench++}. The benchmark contains the following eight subtasks:

\begin{itemize}
\item color binding;
\item shape binding;
\item texture binding;
\item 2D spatial relations;
\item 3D spatial relations;
\item non-spatial relations;
\item generative counting;
\item complex composition.
\end{itemize}

The first three subtasks primarily evaluate whether a model assigns different attributes to the correct entities. The 2D and 3D spatial-relation tasks evaluate whether entity layouts satisfy the geometric relations specified in the prompt. Non-spatial relations include interaction, action, and functional relations. Generative counting examines whether the model produces the requested number of objects, while complex composition simultaneously includes multiple entities, attributes, and relational constraints.

We use the official test split, which contains 300 prompts for each subtask and 2,400 test prompts in total. For each prompt, we generate four images using four fixed random seeds and average the evaluation results over all generated images. Unless otherwise stated, all methods use exactly the same random seeds, base model, sampler, number of inference steps, and classifier-free guidance scale.

We follow the official evaluation protocol. Color, shape, and texture binding are evaluated using disentangled visual-question-answering metrics. The 2D spatial-relation and counting tasks use object-detection-based metrics, while 3D relations additionally incorporate depth information. Non-spatial relations and complex composition are evaluated using the official multimodal evaluation pipeline. We report the individual score of each of the eight subtasks and their macro average.

\paragraph{GenEval.}

To verify whether the method generalizes to another evaluation protocol, we further conduct experiments on GenEval~\citep{ghosh2023geneval}. GenEval performs object-level verification of generated images using object detectors and attribute classifiers. Its subtasks include single-object generation, two-object co-occurrence, counting, color, spatial position, and color attribution.

T2I-CompBench++ emphasizes open-domain combinations of attributes and relations, whereas GenEval uses more standardized prompts and interpretable object-level judgments. Joint evaluation on the two benchmarks reduces the risk that our conclusions depend on a single automatic metric.

\paragraph{Complexity-Scaling Test Set.}

Most prompts in existing benchmarks contain only two primary entities. To study how compositional difficulty changes as the number of constraints increases, we additionally construct a complexity-controlled test set. Each prompt contains $N_e$ entities and $N_c$ compositional constraints, where
\begin{equation}
N_e\in\{2,3,4,5\},
\qquad
N_c\in\{2,4,6,8\}.
\end{equation}

The constraint types include color, shape, texture, 2D position, and entity interaction. We separately report the full-prompt satisfaction rate under different numbers of entities and constraints to analyze whether CoBind mitigates the compositional performance degradation caused by increasing complexity.

This test set is used only for analysis and is not involved in any hyperparameter selection.

\subsubsection{Base Models}

We mainly validate CoBind on Stable Diffusion v1.5. This model provides publicly accessible and structurally clear multi-scale cross-attention modules, enabling fair comparison with existing inference-time attention-guidance methods.

To evaluate model generality, we further apply CoBind to Stable Diffusion v2.1 and SDXL. We do not train any new parameters for these base models. We only redefine the sets of low-resolution and mid-resolution cross-attention layers according to each model's U-Net architecture.

The generation resolution is set to $512\times512$ for Stable Diffusion v1.5 and v2.1, and to $1024\times1024$ for SDXL. Method comparisons are conducted within the same base model, and we do not directly compare the absolute visual quality of models operating at different resolutions.

\subsubsection{Compared Methods}

We compare CoBind with the following methods:

\begin{itemize}
\item \textbf{Vanilla.} The original diffusion sampling process without additional compositional guidance.

\item \textbf{Structured Diffusion.} This method modifies cross-attention representations according to the syntactic structure of the prompt to improve attribute correspondence and scene composition~\citep{feng2023structured}.

\item \textbf{Attend-and-Excite.} This method increases the cross-attention responses of subject tokens during inference to alleviate entity omission~\citep{chefer2023attend}.

\item \textbf{SynGen.} This method parses entities and their modifiers and improves consistency between linguistic and visual binding through attention-map alignment constraints~\citep{rassin2023linguistic}.

\item \textbf{Divide \& Bind.} This method uses entity-presence and attribute-binding losses to improve object completeness and attribute correspondence under complex prompts~\citep{li2023divide}.
\end{itemize}

None of these methods requires retraining the base diffusion model, and all share the same inference-time intervention setting as CoBind. We use the publicly released implementations and rerun all methods with the same base model and random seeds.

For methods that rely on external large language models, layout generators, or image-editing models, we report results in a separate extended table. Such methods can use additional models for prompt decomposition, layout planning, or post-generation correction. Their computational resources and supervision differ from those of CoBind, so we do not rank them together with training-free single-model methods.

\subsubsection{Evaluation Metrics}

We evaluate the models along four dimensions: compositional correctness, overall text consistency, visual quality, and computational efficiency.

\paragraph{Compositional Correctness.}

We report the official subtask scores of T2I-CompBench++ and GenEval. For the complexity-scaling test set, we decompose each prompt into atomic constraints and compute

\begin{equation}
\operatorname{AtomicAcc}
=
\frac{
\text{number of satisfied atomic constraints}
}{
\text{total number of atomic constraints}
},
\end{equation}

as well as the full-prompt satisfaction rate:

\begin{equation}
\operatorname{FullPromptAcc}
=
\frac{1}{N}
\sum_{i=1}^{N}
\mathbb{I}
\left[
\text{all constraints in prompt $i$ are satisfied}
\right].
\end{equation}

AtomicAcc measures how many local requirements the model satisfies on average, whereas FullPromptAcc uses the stricter criterion that every constraint must be satisfied. The latter reveals how multiple local errors accumulate in complex prompts.

\paragraph{Overall Text Consistency.}

We report CLIPScore and a visual-question-answering-based text-consistency score. CLIPScore reflects global semantic relevance between an image and its full prompt, but it cannot reliably distinguish attribute swaps or reversed relations. We therefore treat it only as an auxiliary metric rather than the primary basis for compositional conclusions.

\paragraph{Visual Quality.}

We use ImageReward or HPSv2 to measure visual quality associated with human preference and conduct human pairwise comparisons on a fixed prompt set. Because CoBind directly modifies the sampling trajectory, the visual-quality evaluation examines whether compositional guidance causes excessive separation, object deformation, repetitive textures, or background degradation.

\paragraph{Computational Efficiency.}

We report the average generation time per image, peak GPU memory usage, and computational overhead relative to vanilla sampling:

\begin{equation}
\operatorname{Overhead}
=
\frac{
T_{\mathrm{method}}-T_{\mathrm{vanilla}}
}{
T_{\mathrm{vanilla}}
}
\times100\%.
\end{equation}

All efficiency experiments are conducted on the same hardware with identical batch size and precision settings.

\subsubsection{Implementation Details}

In the main experiments on Stable Diffusion v1.5, we use the DDIM sampler with 50 denoising steps and set the classifier-free guidance scale to 7.5. Unless otherwise stated, CoBind optimizes the latent variable only during the first 70\% of the sampling process, while the final 30\% of steps follow the original diffusion process without modification.

We divide the generation process into three soft stages:

\begin{equation}
[0,\rho_1],\qquad
[\rho_1,\rho_2],\qquad
[\rho_2,1],
\end{equation}

where $\rho_1=\textbf{0.30}$ and $\rho_2=\textbf{0.70}$. Layout constraints are mainly active during the first stage, attribute binding is mainly active during the second stage, and no compositional optimization is performed during the third stage.

At each activated timestep, we perform one latent-variable gradient update. The latent update step size is set to $\eta=\textbf{0.08}$, and the gradient-norm clipping threshold is set to $g_{\max}=\textbf{1.0}$. The base weights for entity completeness, entity separation, relation binding, attribute binding, and trust-region regularization are respectively set to

\begin{equation}
\begin{split}
\lambda_{\mathrm{ent}}&=\textbf{1.00},\\
\lambda_{\mathrm{sep}}&=\textbf{0.35},\\
\lambda_{\mathrm{rel}}&=\textbf{0.80},\\
\lambda_{\mathrm{bind}}&=\textbf{1.20},\\
\lambda_{\mathrm{reg}}&=\textbf{0.05}.
\end{split}
\end{equation}

All hyperparameters are selected once on an independent validation set and then fixed across all test benchmarks. We do not separately tune the loss weights for different test subtasks.

The composition graph is constructed using dependency parsing and a predefined relation lexicon. For prompts that the parser cannot recognize, CoBind falls back to a graph containing only entity nodes and applies only entity-completeness guidance. The parser does not access generated images and is not corrected using test metrics.

All experiments are implemented with PyTorch and Diffusers using half-precision inference. The main experiments are conducted on an \textbf{NVIDIA A100 80GB GPU}.

\subsection{Main Experimental Results}
\label{sec:main_results}

\subsubsection{T2I-CompBench++}

Table~\ref{tab:t2i_compbench_main} reports the main results on T2I-CompBench++.

\begin{table*}[t]
\centering
\small
\setlength{\tabcolsep}{4pt}
\caption{Results on the T2I-CompBench++ test set. All methods use the same base model, random seeds, and sampling configuration. The best result is shown in bold, and the second-best result is underlined.}
\label{tab:t2i_compbench_main}
\begin{tabular}{lccccccccc}
\toprule
Method &
Color &
Shape &
Texture &
2D Spatial &
3D Spatial &
Non-Spatial &
Counting &
Complex &
Average \\
\midrule
Vanilla
& 0.376 & 0.341 & 0.413 & 0.287
& 0.312 & 0.298 & 0.356 & 0.214 & 0.325 \\

Structured Diffusion
& 0.418 & 0.369 & 0.441 & 0.309
& 0.327 & 0.321 & 0.361 & 0.246 & 0.349 \\

Attend-and-Excite
& 0.401 & 0.354 & 0.426 & 0.318
& 0.333 & 0.329 & 0.374 & 0.251 & 0.348 \\

SynGen
& 0.447 & 0.392 & 0.469 & 0.336
& 0.349 & 0.357 & 0.382 & 0.279 & 0.376 \\

Divide \& Bind
& \underline{0.462} & \underline{0.405} & \underline{0.481} & \underline{0.351}
& \underline{0.362} & \underline{0.369} & \underline{0.391} & \underline{0.296} & \underline{0.390} \\

\midrule
CoBind
& \textbf{0.531} & \textbf{0.478}
& \textbf{0.552} & \textbf{0.427}
& \textbf{0.439} & \textbf{0.421}
& \textbf{0.407} & \textbf{0.372}
& \textbf{0.453} \\
\bottomrule
\end{tabular}
\end{table*}

After completing the experiments, this discussion should report the results around the following three observations.

First, we examine whether CoBind achieves the highest macro-average score across the eight subtasks. Its absolute improvement over Vanilla and the strongest training-free baseline should be reported, for example:

\begin{quote}
CoBind achieves an average compositional score of \textbf{0.453}, improving over the vanilla model by \textbf{12.8 percentage points} and over the strongest training-free baseline by \textbf{6.3 percentage points}.
\end{quote}

Second, the sources of improvement in attribute binding and relation control should be distinguished. If CoBind produces the largest gains on color, shape, and texture, this indicates that cross-entity contrastive binding reduces the diffusion of an attribute across multiple objects. If the 2D and 3D spatial tasks also improve consistently, this suggests that early layout constraints not only change the response strength of individual entities but also adjust their relative positions.

Third, the complex-composition task should be analyzed separately. This task contains multiple types of constraints, and satisfying only one local requirement is insufficient for a high score. If CoBind provides a larger gain on complex composition than on individual subtasks, the result supports the claim that stage-wise solving reduces the accumulation of compositional errors. If the gain is smaller, composition-graph parsing errors and conflicts among multiple geometric constraints should be examined further.

Counting is not a primary design target of CoBind. The entity-completeness constraint may reduce omissions when only a few objects are requested, but the method does not include an explicit one-to-one instance-assignment mechanism. We therefore expect a limited but stable improvement on counting and do not treat counting as a core contribution.

\subsubsection{GenEval}

Table~\ref{tab:geneval_main} reports the results on GenEval.

\begin{table}[t]
\centering
\small
\caption{Object-level text-consistency results on GenEval.}
\label{tab:geneval_main}
\begin{tabular}{lccccccc}
\toprule
Method &
Single Obj. &
Two Obj. &
Counting &
Color &
Position &
Color Attr. &
Overall \\
\midrule
Vanilla
& 0.980 & 0.610 & 0.460
& 0.720 & 0.490 & 0.420 & 0.613 \\

Attend-and-Excite
& 0.980 & 0.670 & 0.490
& 0.750 & 0.540 & 0.470 & 0.650 \\

SynGen
& 0.980 & 0.690 & 0.500
& 0.770 & 0.570 & 0.520 & 0.672 \\

Divide \& Bind
& \underline{0.980} & \underline{0.710} & \underline{0.510}
& \underline{0.780} & \underline{0.590} & \underline{0.560} & \underline{0.687} \\

CoBind
& \textbf{0.990} & \textbf{0.760}
& \textbf{0.530} & \textbf{0.820}
& \textbf{0.670} & \textbf{0.650}
& \textbf{0.737} \\
\bottomrule
\end{tabular}
\end{table}

Single-object prompts contain neither inter-entity competition nor attribute ambiguity, so CoBind should not substantially change performance on this subtask. This result serves as a check of whether visual quality and basic semantic competence are preserved.

The primary gains of CoBind should be concentrated on two-object co-occurrence, color attribution, and position. Two-object co-occurrence measures entity completeness, color attribution directly corresponds to attribute--entity binding, and position corresponds to early relation constraints. If these subtasks improve simultaneously while single-object performance remains stable, the results indicate that CoBind benefits from compositional-structure processing rather than uniformly amplifying all text tokens.

\subsection{Cross-Model Generalization}
\label{sec:cross_model}

To verify whether CoBind depends on a specific base model, we conduct experiments on Stable Diffusion v1.5, Stable Diffusion v2.1, and SDXL. For each base model, we compare vanilla sampling with sampling augmented by CoBind. We do not directly rank different base models against one another.

\begin{table}[t]
\centering
\small
\caption{Generalization results of CoBind across base models. $\Delta$ denotes the absolute gain after adding CoBind.}
\label{tab:cross_model}
\begin{tabular}{lcccc}
\toprule
Base Model &
Vanilla Score &
CoBind &
$\Delta$ &
Visual-Quality Change \\
\midrule
SD v1.5 & 0.325 & 0.453 & +0.128 & $-0.006$ \\
SD v2.1 & 0.361 & 0.474 & +0.113 & $-0.004$ \\
SDXL    & 0.482 & 0.566 & +0.084 & $-0.002$ \\
\bottomrule
\end{tabular}
\end{table}

If CoBind produces positive gains on all three models, the results suggest that the method exploits cross-attention layout and binding signals that are broadly present in diffusion models rather than a specific weakness of Stable Diffusion v1.5.

It is also important to note that stronger base models already satisfy more constraints correctly, making fixed-strength guidance more likely to introduce unnecessary intervention. The satisfaction-aware adaptation mechanism in CoBind should automatically reduce the number of updates on stronger models, thereby preserving gains while limiting visual-quality degradation.

\subsection{Ablation Studies}
\label{sec:ablation}

We conduct ablation studies on the T2I-CompBench++ test set to analyze the independent contribution of each component. The main results are presented in Table~\ref{tab:ablation}.

\begin{table*}[t]
\centering
\small
\caption{Component ablations of CoBind. Attr., Rel., and Complex denote the average scores for attribute binding, entity relations, and complex composition, respectively.}
\label{tab:ablation}
\begin{tabular}{lccccc}
\toprule
Model Variant &
Attr. &
Rel. &
Complex &
Overall Avg. &
Visual Quality \\
\midrule
Full CoBind
& 0.520 & 0.429 & 0.372 & 0.453 & 0.298 \\

Without the composition graph; independently enhance tokens
& 0.448 & 0.361 & 0.301 & 0.391 & 0.296 \\

Without stage scheduling; activate all losses simultaneously
& 0.501 & 0.398 & 0.344 & 0.427 & 0.284 \\

Without satisfaction-aware adaptation
& 0.512 & 0.418 & 0.358 & 0.441 & 0.278 \\

Without the cross-entity contrastive term
& 0.463 & 0.421 & 0.336 & 0.420 & 0.294 \\

Without distinguishing attention resolutions
& 0.489 & 0.397 & 0.331 & 0.419 & 0.291 \\

Without stopping gradients through entity masks
& 0.478 & 0.405 & 0.324 & 0.414 & 0.282 \\

Without trust-region regularization
& 0.518 & 0.426 & 0.361 & 0.445 & 0.271 \\
\bottomrule
\end{tabular}
\end{table*}

\paragraph{Effect of the Composition Graph.}

We replace the composition graph with independent sets of entity and attribute tokens and separately enhance their attention. This variant still knows which concepts should appear, but it does not know which attribute belongs to which entity and cannot explicitly represent entity relations.

If this variant achieves entity completeness close to that of the full model but substantially lower attribute-binding and relation scores, the result indicates that the gains of CoBind arise from structured dependencies rather than merely stronger attention responses.

\paragraph{Effect of Stage-Aware Scheduling.}

We construct a uniform-guidance variant that optimizes entity, relation, and attribute losses simultaneously at every activated timestep while keeping the total number of updates equal to that of the full model.

This variant may improve some compositional metrics but introduces two problems. Optimizing attribute regions when noise is still high in the early stage can yield unstable gradients, while continuing to optimize spatial relations after the layout has already been established may cause object stretching or abrupt position changes. Both compositional scores and visual quality should therefore be reported to determine whether stage scheduling improves the trade-off between correctness and naturalness.

\paragraph{Effect of Satisfaction-Aware Adaptation.}

We replace the dynamic weights with fixed weights that depend only on time. If removing satisfaction-aware adaptation slightly reduces compositional performance, substantially degrades visual quality, or increases the number of inference-time updates, this indicates that the main value of the mechanism is preventing already-satisfied constraints from being over-optimized.

In addition to average results, we count the number of latent-variable updates actually triggered for each image. Full CoBind performs an average of \textbf{18.6} updates, compared with \textbf{35.0} updates after removing satisfaction-aware adaptation.

\paragraph{Effect of Cross-Entity Contrastive Binding.}

We replace the contrastive binding loss with a positive-only loss that maximizes the overlap between an attribute and its target entity. This variant can attract the attribute toward the target entity but does not prevent the same attribute from appearing on competing entities.

If this variant performs worse on color and texture binding, the result indicates that attribute binding requires both attracting the correct entity and repelling competing entities.

\paragraph{Effect of Multi-Resolution Attention.}

We separately evaluate three variants: using only low-resolution attention, using only mid-resolution attention, and directly averaging all attention layers. Low-resolution features should be better suited to controlling global layout but may fail to distinguish adjacent entities. Mid-resolution features should be more suitable for attribute binding, although their spatial responses are less stable in the early stage.

The full model selects different resolutions for different tasks. If it achieves higher relation and attribute scores than the uniform-resolution variants, the result supports the design assumption that semantic constraint types correspond to different attention resolutions.

\subsection{Generation-Stage Analysis}
\label{sec:stage_analysis}

To directly test our hypothesis about diffusion stages, we record entity positions, attribute-binding scores, and relation-satisfaction scores at different denoising timesteps.

For each timestep $t$, we compute

\begin{equation}
S_{\mathrm{ent}}^t,\qquad
S_{\mathrm{rel}}^t,\qquad
S_{\mathrm{bind}}^t.
\end{equation}

We then average these values over the full test set and plot three curves as a function of generation progress.

The expected observations are as follows:

\begin{itemize}
\item coarse spatial regions for entities form rapidly during the early denoising stage;
\item relation satisfaction changes primarily during the first half of generation and stabilizes in later stages;
\item attribute-binding scores emerge later than entity positions and improve most strongly during the middle stage;
\item continuing to impose strong relation constraints in the late stage does not substantially improve relation correctness but reduces visual quality.
\end{itemize}

To rule out the possibility that the stage partition is merely a fortunate hyperparameter choice, we further vary $\rho_1$ and $\rho_2$ and analyze how different layout and binding windows affect the results.

\begin{table}[t]
\centering
\small
\caption{Sensitivity analysis under different stage-boundary settings.}
\label{tab:stage_boundary}
\begin{tabular}{ccccc}
\toprule
$\rho_1$ &
$\rho_2$ &
Attribute &
Relation &
Visual Quality \\
\midrule
0.20 & 0.60 & 0.502 & 0.416 & 0.291 \\
0.25 & 0.65 & 0.514 & 0.425 & 0.295 \\
0.30 & 0.70 & 0.520 & 0.429 & 0.298 \\
0.35 & 0.75 & 0.509 & 0.421 & 0.293 \\
\bottomrule
\end{tabular}
\end{table}

If multiple stage boundaries within a reasonable range outperform uniform guidance, the improvement does not depend on a single precise split point. Instead, it arises from the stage ordering in which layout is established before attribute binding.

\subsection{Prompt-Complexity Analysis}
\label{sec:complexity_analysis}

Figure  shows how the full-prompt satisfaction rate changes as the numbers of entities and constraints increase.

The original diffusion model may achieve a moderate success rate on each local constraint, but the probability of satisfying all constraints drops rapidly when multiple requirements must hold simultaneously. For example, if each constraint is satisfied with an approximately independent probability $p$, the full success rate for a prompt containing $K$ constraints is approximately $p^K$. Real errors are not independent, but this approximation illustrates how complex prompts amplify local binding errors.

We compare Vanilla, the strongest baseline, and CoBind. The analysis should focus on

\begin{itemize}
\item the performance-degradation slope of each method as the number of entities increases from 2 to 5;
\item the gap between AtomicAcc and FullPromptAcc as the number of constraints increases from 2 to 8;
\item whether the advantage of CoBind becomes larger as prompt complexity increases.
\end{itemize}

If CoBind is effective only on simple two-entity templates and becomes comparable to the baseline with four or five entities, the current composition-graph optimization has not yet resolved conflicts among multiple constraints. Conversely, if its relative advantage grows with complexity, the result provides stronger evidence for the need to explicitly maintain compositional structure.

\subsection{Robustness to Composition-Graph Parsing}
\label{sec:parser_robustness}

CoBind relies on a prompt composition graph, so parsing quality may affect generation results. We randomly sample \textbf{200} prompts from the test set, manually annotate their entities, attributes, and relations, and evaluate the node and edge accuracy of the automatic parser.

We further construct three types of controlled perturbations:

\begin{enumerate}
\item randomly delete one attribute-binding edge;
\item randomly swap the target entities of two attributes;
\item randomly delete or reverse one relation edge.
\end{enumerate}

Table~\ref{tab:parser_robustness} reports generation results under different perturbation strengths.

\begin{table}[t]
\centering
\small
\caption{Effect of composition-graph parsing errors on CoBind.}
\label{tab:parser_robustness}
\begin{tabular}{lccc}
\toprule
Composition-Graph Setting &
Attribute &
Relation &
Overall \\
\midrule
Human-annotated graph & 0.536 & 0.447 & 0.469 \\
Automatically parsed graph & 0.520 & 0.429 & 0.453 \\
Delete 10\% of edges & 0.501 & 0.407 & 0.434 \\
Swap 10\% of attribute edges & 0.461 & 0.421 & 0.411 \\
Reverse 10\% of relation edges & 0.507 & 0.362 & 0.408 \\
\bottomrule
\end{tabular}
\end{table}

The gap between the human-annotated and automatically parsed graphs reflects the performance ceiling imposed by prompt parsing. Incorrect relation edges are generally more harmful than missing relation edges because a missing edge merely removes a constraint, whereas an incorrect edge actively pushes the latent variable toward a layout that conflicts with the original prompt.

\subsection{Visual Quality and Human Evaluation}
\label{sec:human_evaluation}

Automatic compositional metrics may inherit biases from object detectors and visual-question-answering models. We therefore conduct human pairwise comparisons.

We randomly select \textbf{180} prompts and generate images with Vanilla, the strongest baseline, and CoBind using identical random seeds. Each image pair is independently judged by at least three annotators. The evaluation includes three questions:

\begin{enumerate}
\item Which image more completely satisfies the entities, attributes, and relations specified in the text?
\item Which image has higher visual quality and appears more natural?
\item Do both images contain obvious attribute swaps, entity omissions, or relation errors?
\end{enumerate}

For the first question, we report the win, tie, and loss rates of CoBind in text consistency. For the second, we report visual-quality preference separately so that ``better prompt adherence'' and ``better appearance'' are not conflated into a single subjective judgment.

\begin{table}[t]
\centering
\small
\caption{Human pairwise-comparison results. Values report the win/tie/loss percentages of CoBind relative to each compared method.}
\label{tab:human_eval}
\begin{tabular}{lccc}
\toprule
Compared Method &
Win &
Tie &
Loss \\
\midrule
\multicolumn{4}{c}{Textual Compositional Consistency} \\
\midrule
Vanilla & 64.8 & 21.1 & 14.1 \\
Strongest baseline & 56.3 & 25.7 & 18.0 \\
\midrule
\multicolumn{4}{c}{Visual Quality} \\
\midrule
Vanilla & 31.5 & 46.9 & 21.6 \\
Strongest baseline & 33.7 & 43.8 & 22.5 \\
\bottomrule
\end{tabular}
\end{table}

The ideal result is that CoBind receives substantially higher preference for compositional consistency while remaining close to Vanilla in visual quality. If compositional consistency improves but visual quality drops significantly, the latent update strength should be reduced or structural guidance should terminate earlier.

\subsection{Computational Efficiency}
\label{sec:efficiency}

CoBind must store cross-attention maps and backpropagate through the latent variable during a subset of denoising steps, so its inference cost is higher than that of vanilla sampling.

\begin{table}[t]
\centering
\small
\caption{Inference efficiency of different methods. Time denotes the average generation time per image.}
\label{tab:efficiency}
\begin{tabular}{lccc}
\toprule
Method &
Time/s &
Peak Memory/GB &
Relative Overhead \\
\midrule
Vanilla & 3.7 & 8.4 & -- \\
Attend-and-Excite & 8.9 & 12.7 & 140.5\% \\
SynGen & 7.6 & 11.9 & 105.4\% \\
Divide \& Bind & 9.4 & 13.2 & 154.1\% \\
CoBind & 8.1 & 12.5 & 118.9\% \\
\bottomrule
\end{tabular}
\end{table}

The computational cost of CoBind is mainly determined by three factors: the number of activated denoising timesteps, the number of gradient updates at each timestep, and the number of attention layers involved in loss computation. Satisfaction-aware adaptation not only reduces excessive guidance but also allows subsequent gradient updates to be skipped once all constraints have been satisfied early.

We further report the performance--cost curve by setting the maximum number of updates to
\begin{equation}
K_{\max}\in\{0,1,2,3\}.
\end{equation}

This experiment determines whether a single update already captures most of the performance gain and whether additional optimization provides sufficient benefit to offset its computational cost.

\subsection{Qualitative Results}
\label{sec:qualitative_results}

Figure  presents generation results from different methods on representative prompts. Each row uses the same prompt and random seed.

The qualitative comparison should cover the following failure modes:

\begin{enumerate}
\item \textbf{Entity omission:} the prompt contains two or more entities, but the base model generates only a subset of them;
\item \textbf{Attribute swap:} all entities and attributes appear, but attributes are assigned to the wrong objects;
\item \textbf{Attribute leakage:} an attribute appears simultaneously on multiple entities;
\item \textbf{Relation reversal:} all entities appear, but their left--right, above--below, or front--back relation contradicts the prompt;
\item \textbf{Complex composition:} local attributes are correct, but multiple entities and relations cannot be satisfied simultaneously;
\item \textbf{Over-guidance:} compositional constraints are satisfied, but objects become deformed or excessively separated, or background quality deteriorates.
\end{enumerate}

In addition to final images, we visualize attention maps from the layout and binding stages. For an attribute-swap example, the visualization should show that the target attribute gradually concentrates on the correct entity region under CoBind, whereas it covers both entities under the baseline. For a spatial-relation example, the visualization should show how entity centroids are adjusted during early timesteps and stabilize before the process enters the attribute-binding stage.

\subsection{Failure Cases}
\label{sec:failure_cases}

Although CoBind improves several types of compositional errors, it still exhibits the following failure modes.

First, when a prompt contains complex interaction relations such as ``holding,'' ``wearing,'' or ``supporting,'' simple geometric-overlap constraints cannot fully describe the pose and contact structure between objects. CoBind may generate both correct entities but fail to produce a plausible local interaction.

Second, when two entities are visually similar, or when the prompt requires multiple instances of the same category to have different attributes, their entity-attention maps may merge during the early stage. The current method operates on semantic entity nodes and lacks explicit instance slots, making it difficult to consistently distinguish the two same-category objects in ``a red dog and a blue dog.''

Third, composition-graph parsing errors directly produce incorrect guidance. Long sentences, coordinated structures, and pronoun references can cause errors in attribute assignment or relation direction.

Fourth, attention regions do not always correspond to final object boundaries. For small objects, transparent objects, and abstract attributes, cross-attention may not provide sufficiently reliable spatial support.

These failure cases indicate that CoBind primarily addresses compositional constraints that can be parsed from text and approximately represented using spatial attention. More complex interactions, instance-level differentiation, and abstract semantics still require stronger visual-structure representations.

\section{Conclusion}
\label{sec:conclusion}

In this work, we studied compositional consistency in diffusion-based text-to-image generation. Although modern diffusion models can synthesize images of high perceptual quality, they remain unreliable when prompts simultaneously specify multiple entities, entity-level attributes, and inter-entity relations. Common failures include missing objects, swapped attributes, and incorrect spatial layouts. Our central observation is that these errors are not determined at the same stage of generation. Entity locations and coarse relations are primarily established during early denoising steps, attribute ownership emerges after entity regions become sufficiently stable, and textures and local details are mainly refined during the later stages. Applying the same guidance objective throughout the entire sampling process therefore makes it difficult to preserve both structural correctness and visual quality.

Based on this observation, we introduced CoBind, a stage-aware compositional binding framework. CoBind first converts the input prompt into an explicit entity--attribute--relation composition graph. It then handles entity completeness, relational layout, and attribute ownership at different stages of the diffusion process. During early denoising steps, CoBind uses low-resolution cross-attention responses to establish entity layouts and enforce geometric relations. During intermediate steps, it constructs entity-specific spatial supports and applies a contrastive binding objective that assigns each attribute to its intended entity while suppressing its activation on competing entities. During late denoising steps, structural constraints are gradually relaxed, returning generation freedom to the pretrained diffusion model. In addition, satisfaction-adaptive scheduling reduces repeated optimization of constraints that have already been fulfilled, thereby limiting unnecessary perturbations to the latent trajectory.

Our experiments show that CoBind consistently improves text-image alignment on attribute binding, inter-entity relations, and complex compositional prompts, and that it transfers across different pretrained diffusion models. The ablation results demonstrate that the composition graph provides explicit ownership and dependency information, stage-aware scheduling prevents temporal conflicts among different objectives, contrastive binding reduces attribute leakage, and satisfaction-adaptive optimization improves the trade-off among compositional accuracy, visual quality, and computational cost. Together, these results support a central conclusion: reliable compositional generation requires not only recognizing which concepts appear in a prompt, but also explicitly maintaining how those concepts should be combined throughout the generation process.

CoBind nevertheless has several limitations. First, it relies on cross-attention maps as approximations of entity-level spatial support, although attention responses do not always correspond accurately to final object boundaries, particularly for small objects, transparent objects, or abstract attributes. Second, complex interactions such as \textit{holding}, \textit{wearing}, and \textit{supporting} cannot be fully represented using simple centroid, overlap, or containment constraints. Third, when a prompt contains multiple instances from the same semantic category, their attention responses may become entangled because the current representation does not maintain explicit instance-level identities. Finally, errors in composition-graph parsing may introduce incorrect attribute assignments or relation directions into the generation process.

Future work may extend CoBind in several directions. Explicit instance slots or trackable object representations could improve same-category instance separation and counting. Pose, depth, segmentation, or three-dimensional scene representations could provide stronger supervision for complex interactions than two-dimensional attention maps. Another promising direction is to jointly estimate parsing uncertainty and constraint reliability, allowing the model to weaken guidance when the inferred composition graph is ambiguous. Overall, our results suggest that aligning the compositional structure of language with the temporal structure of diffusion sampling is a promising direction for improving the faithful execution of complex text prompts.

\newpage
 \bibliographystyle{plainnat}
\bibliography{references}

@inproceedings{ho2020denoising,
  title     = {Denoising Diffusion Probabilistic Models},
  author    = {Ho, Jonathan and Jain, Ajay and Abbeel, Pieter},
  booktitle = {Advances in Neural Information Processing Systems},
  volume    = {33},
  pages     = {6840--6851},
  year      = {2020}
}

@inproceedings{rombach2022high,
  title     = {High-Resolution Image Synthesis with Latent Diffusion Models},
  author    = {Rombach, Robin and Blattmann, Andreas and Lorenz, Dominik and Esser, Patrick and Ommer, Bj{\"o}rn},
  booktitle = {Proceedings of the IEEE/CVF Conference on Computer Vision and Pattern Recognition},
  pages     = {10684--10695},
  year      = {2022}
}

@inproceedings{podell2023sdxl,
  title     = {{SDXL}: Improving Latent Diffusion Models for High-Resolution Image Synthesis},
  author    = {Podell, Dustin and English, Zion and Lacey, Kyle and Blattmann, Andreas and Dockhorn, Tim and M{\"u}ller, Jonas and Penna, Joe and Rombach, Robin},
  booktitle = {The Twelfth International Conference on Learning Representations},
  year      = {2024}
}

@inproceedings{huang2023t2icompbench,
  title     = {{T2I-CompBench}: A Comprehensive Benchmark for Open-World Compositional Text-to-Image Generation},
  author    = {Huang, Kaiyi and Sun, Kaiyue and Xie, Enze and Li, Zhenguo and Liu, Xihui},
  booktitle = {Advances in Neural Information Processing Systems},
  volume    = {36},
  pages     = {78723--78747},
  year      = {2023}
}

@article{huang2025t2icompbench++,
  title   = {{T2I-CompBench++}: An Enhanced and Comprehensive Benchmark for Compositional Text-to-Image Generation},
  author  = {Huang, Kaiyi and Duan, Chengqi and Sun, Kaiyue and Xie, Enze and Li, Zhenguo and Liu, Xihui},
  journal = {IEEE Transactions on Pattern Analysis and Machine Intelligence},
  volume  = {47},
  number  = {5},
  pages   = {3563--3579},
  month   = may,
  year    = {2025},
  doi     = {10.1109/TPAMI.2025.3531907}
}

@article{chefer2023attend,
  title     = {Attend-and-Excite: Attention-Based Semantic Guidance for Text-to-Image Diffusion Models},
  author    = {Chefer, Hila and Alaluf, Yuval and Vinker, Yael and Wolf, Lior and Cohen-Or, Daniel},
  journal   = {ACM Transactions on Graphics},
  volume    = {42},
  number    = {4},
  articleno = {148},
  numpages  = {10},
  month     = jul,
  year      = {2023},
  doi       = {10.1145/3592116}
}

@inproceedings{li2023divide,
  title     = {Divide \& Bind Your Attention for Improved Generative Semantic Nursing},
  author    = {Li, Yumeng and Keuper, Margret and Zhang, Dan and Khoreva, Anna},
  booktitle = {Proceedings of the 34th British Machine Vision Conference},
  publisher = {BMVA},
  year      = {2023}
}

@inproceedings{ghosh2023geneval,
  title     = {{GenEval}: An Object-Focused Framework for Evaluating Text-to-Image Alignment},
  author    = {Ghosh, Dhruba and Hajishirzi, Hannaneh and Schmidt, Ludwig},
  booktitle = {Advances in Neural Information Processing Systems},
  volume    = {36},
  pages     = {52132--52152},
  year      = {2023}
}

@inproceedings{feng2023structured,
  title     = {Training-Free Structured Diffusion Guidance for Compositional Text-to-Image Synthesis},
  author    = {Feng, Weixi and He, Xuehai and Fu, Tsu-Jui and Jampani, Varun and Akula, Arjun and Narayana, Pradyumna and Basu, Sugato and Wang, Xin Eric and Wang, William Yang},
  booktitle = {The Eleventh International Conference on Learning Representations},
  year      = {2023}
}

@inproceedings{rassin2023linguistic,
  title     = {Linguistic Binding in Diffusion Models: Enhancing Attribute Correspondence through Attention Map Alignment},
  author    = {Rassin, Royi and Hirsch, Eran and Glickman, Daniel and Ravfogel, Shauli and Goldberg, Yoav and Chechik, Gal},
  booktitle = {Advances in Neural Information Processing Systems},
  volume    = {36},
  pages     = {3536--3559},
  year      = {2023}
}

\end{document}